\documentclass[sigconf]{acmart}

\AtBeginDocument{%
  \providecommand\BibTeX{{%
    \normalfont B\kern-0.5em{\scshape i\kern-0.25em b}\kern-0.8em\TeX}}}

\setcopyright{acmlicensed}
\copyrightyear{2024}
\acmYear{2024}
\acmDOI{XXXXXXX.XXXXXXX}

\acmISBN{978-1-4503-XXXX-X/18/06}

\usepackage{fancybox}
\begin{document}

\title[Quantifying Geospatial in the Common Crawl Corpus]{Quantifying Geospatial in the Common Crawl Corpus}

\author{Ilya Ilyankou}
\authornote{Corresponding author.}
\email{ilya.ilyankou.23@ucl.ac.uk}
\orcid{0009-0008-7082-7122}

\author{Meihui Wang}
\email{meihui.wang.20@ucl.ac.uk}
\orcid{0000-0001-8420-2141}

\affiliation{%
  \institution{UCL SpaceTimeLab}
  \city{London}
  \country{UK}
}

\author{Stefano Cavazzi}
\email{stefano.cavazzi@os.uk}
\orcid{0000-0003-3575-0365}
\affiliation{%
  \institution{Ordnance Survey}
  \city{Southampton}
  \country{UK}
}

\author{James Haworth}
\email{j.haworth@ucl.ac.uk}
\orcid{0000-0001-9506-4266}

\affiliation{%
  \institution{UCL SpaceTimeLab}
  \city{London}
  \country{UK}
}


\begin{abstract}
Large language models (LLMs) exhibit emerging geospatial capabilities, stemming from their pre-training on vast unlabelled text datasets that are often derived from the Common Crawl (CC) corpus. However, the geospatial content within CC remains largely unexplored, impacting our understanding of LLMs' spatial reasoning. This paper investigates the prevalence of geospatial data in recent Common Crawl releases using Gemini 1.5, a powerful language model. By analyzing a sample of documents and manually revising the results, we estimate that 18.7\% of web documents in CC contain geospatial information such as coordinates and addresses. We find little difference in prevalence between Enlgish- and non-English-language documents. Our findings provide quantitative insights into the nature and extent of geospatial data in CC, and lay the groundwork for future studies of geospatial biases of LLMs.
\end{abstract}



\received{31 May 2024}

\maketitle

\section{Introduction}

Recent studies have found that large language models (LLMs) may have some knowledge and understanding of physical space. Gurnee and Tegmark \cite{gurnee_language_2024} demonstrated that Llama-2 learned linear representations of space at various scales, such as landmarks and cities; the authors even identified specific nodes in the model's architecture that are responsible for this behaviour. Bhandari et al. \cite{bhandari_are_2023} probed Llama and several smaller language models for geocoding abilities and geospatial awareness, showing that larger and more sophisticated models have better abilities to synthesise geospatial knowledge from pre-training texts. Mai et al. \cite{mai_opportunities_2023} demonstrated superior performance of the GPT-family LLMs on toponym recognition and location description recognition tasks, with few-shot learning often performing on par with the specialised, state-of-the-art named-entity recognition (NER) models.

It is generally accepted that language models acquire most of their knowledge during the unsupervised pre-training process, when large amounts of unlabelled and unstructured text are fed into the model \cite{radford_improving_2018}. The Common Crawl corpus\footnote{\href{https://commoncrawl.org/}{https://commoncrawl.org/}} (CC) is the largest free web crawl dataset
that powers many smaller, higher-quality datasets such as OSCAR\footnote{\href{https://oscar-project.org/}{https://oscar-project.org/}} \cite{suarez_asynchronous_2021}, CC-100\footnote{\href{https://huggingface.co/datasets/cc100}{https://huggingface.co/datasets/cc100}}, Colossal Cleaned Crawl Corpus (C4)\footnote{\href{https://huggingface.co/datasets/c4}{https://huggingface.co/datasets/c4}} \cite{raffel_exploring_2020}, and RefinedWeb\footnote{\href{https://huggingface.co/datasets/tiiuae/falcon-refinedweb}{https://huggingface.co/datasets/tiiuae/falcon-refinedweb}} \cite{penedo_refinedweb_2023}, which are often used in pre-training of LLMs.

Given the importance of CC in the field of generative AI and the growing concerns over the biases and ethics of the generative AI models, it is not surprising the corpus has been analysed for undesirable content such as hate speech and sexually explicit content \cite{luccioni_whats_2021, raffel_exploring_2020}, and its derivative datasets were analysed for general quality \cite{kreutzer_quality_2022, dodge_documenting_2021, birhane_multimodal_2021}. While the geospatial bias of LLMs has been subject to multiple investigations \cite{moayeri_worldbench_2024, manvi_large_2024, fulman_evidence_2024}, sometimes producing disturbing results (such as that the `error rates are 1.5 times higher for countries from Sub-Saharan Africa compared to North American countries' \cite{moayeri_worldbench_2024} and `the LLMs exhibit geographic bias across objective and subjective topics, particularly discriminating against areas with lower socioeconomic conditions' \cite{manvi_large_2024}), to our knowledge, no attempts were made to look into the geospatial contents of CC.

In this paper, we aim to fill in this gap by quantifying the number of geospatial files found in recent CC releases, as well as estimating the share of traditional web documents (for example, HTML and XML files) that contain geospatial information, such as coordinates or street addresses. We believe that any debate about the emergent geospatial capabilities of LLMs, as well as the geospatial biases they demonstrate, cannot be complete without knowing the prevalence of geospatial content in the pre-training text corpus.

In the following section, we will present an overview of existing literature quantifying various types of content in CC and the derived datasets. We will then suggest our method of quantifying geospatial content in CC, and present our findings based on three CC releases spaced out over the last five years.

\section{Related work}

In 2012, Mühleisen and Bizer \cite{muhleisen_web_2012} suggested a pipeline to extract structured data, such as Microformats\footnote{https://developer.mozilla.org/en-US/docs/Web/HTML/microformats}, RDFa\footnote{https://www.w3.org/TR/rdfa-primer/} and Microdata\footnote{https://developer.mozilla.org/en-US/docs/Web/HTML/Microdata}, from CC. The authors estimated that 6\% of the HTML pages found in the 2009/10 version of CC contained at least one type of structured data, increasing to 12\% in the February 2012 version. Such a jump is not surprising given the exploding popularity of structured data on the web since their creation in mid-2000s \cite{ruth_linked_2013}. According to \cite{muhleisen_web_2012}, only about 2\% of RDFa and 8\% of Microdata objects contained geographical information.

To our knowledge, no other research focused on quantifying geospatial in CC and its derivative datasets. When it comes to other types of content, in 2021, Dodge et al. \cite{dodge_documenting_2021} claimed to be some of the first researchers to analyse the C4 dataset, which is a subset of CC \cite{raffel_exploring_2020}. They identified a large number of patents, documents coming from the US military websites, and machine-generated text, and concluded that the text from and about minority individuals that is present in CC is disproportionately removed from C4. In the same year, Luccioni and Viviano \cite{luccioni_whats_2021} randomly sampled 1\% of the documents in a single CC release, and applied n-grams, HateSonar\footnote{https://github.com/Hironsan/HateSonar}, and DE-LIMIT\footnote{https://github.com/hate-alert/DE-LIMIT} methods to estimate the prevalence of hate speech in sampled documents at 4.02\% to 6.38\%, and the prevalence of sexually explicit content at between 0.73\% and 2.36\%. In a somewhat peculiar exploration in 2014, van Hague et al. \cite{van_hage_number_2014} identified the frequency of numbers found in the 2012 version of CC, showing that the (overwhelmingly positive) numbers found across web pages do follow the Power law distribution.

These works gave us insight into how researchers approach the problem of analysing and quantifying phenomena in large text corpora. We propose our own method of quantifying geospatial in CC in the following section.

\section{Method}

Our aim is to estimate what \emph{share} of documents in a typical CC release contains \emph{any} geospatial information as defined in Subsection \ref{WhatConstituesGeospatial}. We deliberately avoid labelling and then quantifying the \emph{actual} geospatial text as our delineation criteria and judgement of relevance would get extremely complex.

\subsection{About Common Crawl}

Common Crawl is a San-Francisco based non-profit organisation. Established in 2007, its goal is to `make wholesale extraction, transformation, and analysis of open web data accessible to researchers' \cite{common_crawl_common_2024}. The released datasets comprise the world's largest free, multi-lingual corpus of scraped web data \cite{baack_training_2024}, estimated to contain about 250 billion pages (over 9.5 petabytes), with between 3 and 5 billion pages added each month \cite{common_crawl_common_2024}. Due to the dataset's scale, it is sometimes incorrectly described as containing `the entire web'. As Baack \cite{baack_training_2024} points out, it is not even a representative sample of the web.

New CC releases are published every one-two months, and are freely available via S3 and HTTP protocols\footnote{https://data.commoncrawl.org}. Each release is divided into 300 Web Archive (WARC) chunks for easier access, and accompanied by the index with relevant metadata on each web document, such as the web document's original URL, MIME-type, and content length. One can obtain \emph{individual} web documents from the WARC chunks directly from the Common Crawl servers, as we demonstrated in our recent paper \cite{ilyankou_cc-gpx_2024}:

\begin{verbatim}
requests.get(
    f'https://data.commoncrawl.org/{warc_file}'
    headers={
        'Range': f'bytes={offset}-{offset+length-1}'
    }
)
\end{verbatim}

In the code snippet above, we use \texttt{requests} Python library\footnote{https://docs.python-requests.org/en/latest/index.html} to retrieve an individual scraped web document from a CC \texttt{warc\_file} using the \texttt{Range} HTTP header, only reading the bytes of the relevant web document by knowing its location within the \texttt{warc\_file}, including the \texttt{offset} from the start of the \texttt{warc\_file}, and the web document's \texttt{length} in bytes.

In this work, we will focus on three CC releases spanning the last five years, \texttt{CC-MAIN-\{2019-09,2021-39,2024-10\}}.

\subsection{What constitutes geospatial} \label{WhatConstituesGeospatial}

The vast majority of the files found in CC are standard web pages, such as HTML and XML pages. In fact, the share of CC files that are exclusively geospatial, such as .geojson or .kml, is only roughly 0.003-0.004\%. Table \ref{tab:mime_geo} shows all geospatial MIME types that we were able to identify from the index of the three studied CC releases. Therefore, our main goal is to estimate the prevalence of geospatial content in predominantly non-geospatial files.

\begin{table}
  \caption{Exclusively geospatial files found in selected CC releases. MIME-types for `application' are abbreviated as `app.'}
  \label{tab:mime_geo}
  \begin{tabular}{lrrr}
    \toprule
     & \multicolumn{3}{c}{CC Release} \\
    MIME file type & 2019-09 & 2021-39 & 2024-10 \\
    \midrule
    app./geo+json & 0 & 11 & 1 \\
    app./gml+xml & 232 & 80 & 333 \\
    app./gpx+xml & 6,071 & 13,144 & 16,951 \\
    app./tcx+xml & 12 & 0 & 0 \\
    app./vnd.garmin.gpx+xml & 1 & 0 & 1 \\
    app./vnd.garmin.tcx+xml & 4 & 0 & 0 \\
    app./vnd.geo+json & 1 & 0 & 0 \\
    app./vnd.google-earth.gpx+xml & 0 & 16 & 57 \\
    app./vnd.google-earth.kml+xml & 67,982 & 67,797 & 108,563 \\
    app./vnd.google-earth.kmz & 14,870 & 12,849 & 12,124 \\
    app./vnd.openstreetmap.data+xml & 0 & 172 & 70 \\
    app./vnd.topografix.gpx+xml & 2 & 0 & 49 \\
    app./x-shapefile & 512 & 219 & 961 \\
    text/gpx+xml & 2 & 2 & 15 \\
    text/x-gpx+xml & 0 & 0 & 1 \\
    text/xml+georss & 4 & 0 & 0 \\
    \midrule
    Total geospatial files & 89,693 & 94,290 & 139,126 \\
    \midrule
    All documents in release & 2914M & 2930M & 3107M \\
    \midrule
    Geo. files as \% of all docs. & 0.003 & 0.003 & 0.004 \\
  \bottomrule
\end{tabular}
\end{table}


According to the ISO's definition of \emph{geographic} data (which is essentially the same as \emph{geospatial} data), it must contain `implicit or explicit reference to a location relative to the Earth' \cite{iso_geographic_2012}.  
In the \emph{Manual of Geospatial Science and Technology}, Bossler defines geospatial data as `features tied to the surface of the earth by coordinates, addresses, or other means' \cite{bossler_manual_2010}. In this paper, we focus on the first two reference types:

\begin{itemize}
    \item \emph{coordinates}, such as latitude-longitude pairs or easting-northing of the British National Grid, whether or not they are accompanied by an address or a location description, and
    \item \emph{street addresses}, specific enough so that they can be geocoded into a point.
\end{itemize}

While we agree that there exist numerous other, lesser used but perfectly valid ways to express locations precise enough to pinpoint a building or a business, such as \emph{`the cafe at the foot of the Eiffel tower'} or descriptors that use local landmarks and unique characteristics to identify places within informal settlements (references which Bossler may have classed as \emph{other means}), after many discussions, we decided not to include them in our calculations for two reasons. Firstly, the threshold for what to include would become fuzzy. Secondly, we would be reliant on the LLM's perception of spatial reasoning and understanding of cultural contexts and navigation to identify and return these examples, inevitably introducing bias to our results.

We do not count derivative location systems such as \texttt{what3words}\footnote{\href{https://what3words.com}{https://what3words.com}} or \texttt{geohash}\footnote{\href{https://geohash.org/}{https://geohash.org/}} as valid for our purposes. (In real life, these are typically accompanied by traditional addresses or coordinates, so we are likely to still count the documents.) We exclude simple mentions of landmarks, such as `the White House', unless they are accompanied by an address (e.g., `The White House, 1600 Pennsylvania Avenue NW, Washington, DC' or even simply `The White House, Washington DC'). This is to ensure we don't count documents that mention landmarks in passing, although we recognise it can be argued that LLMs may learn the physical world from such examples as well.

We do not discriminate against real and fictional addresses. For example, we count `123 Main Street<br/>New York, NY 10001' as a valid address even though we are certain it is used as a filler. Similarly, we do not check if latitude-longitude pairs are valid as long as it is clear the pair of numbers represents physical coordinates. We do, however, exclude the coordinates of the Null Island (0,0) as those are often used for map initialisation and are not meaningful otherwise \cite{juhasz_i_2022}. We also exclude astronomical coordinates of celestial bodies where possible.

\subsection{Identifying coordinates and addresses} \label{IdentifyingCoordinatesAndAddresses}

In web documents, the coordinates most commonly appear inside Google Maps links, inside JavaScript code snippets for map initialisation or marker locations, or as part of \texttt{<meta>} tags or structured documents (e.g., Schema.org\footnote{https://schema.org/}):

\begin{verbatim}
https://maps.google.co.jp/maps?q=35.6655165,139.7284844
                        —
new google.maps.LatLng(38.674594,15.889066)
                        —
GPS:<br />49°12'9.862"N<br />18°45'19.909"E
                        —
<meta name="geo.position" content="48.85837;2.294481" />
                        —
{
  "@type": "GeoCoordinates",
  "latitude": "35.935775",
  "longitude": "-75.612923"
}
                        —
{
  "spatialReference": {
    "wkid": 27700
  },
  "x": 263779.76690000016,
  "y": 235218.1986999996
}
\end{verbatim}

In general, identifying coordinates can be performed using regular expressions, as those are typically expressed as pairs of numbers. For example, latitudes and longitudes are expressed as real numbers between $[-90.0, 90.0]$ and $[-180.0; 180.0]$, respectively. However, any regex pattern is likely to produce false positive matches, such as scientific numbers ($\pi\approx3.141592$) or unusually formatted phone numbers (e.g., $+1.234567890$). In addition, we occasionally see coordinates expressed in minutes and seconds, such as `N 56° 59' 55'' E 009° 19' 33,7''', as well as those expressed using various national grid systems, such as positive integers representing easting and northing in the British National Grid\footnote{https://britishnationalgrid.uk/}. Adding more regular expressions to capture all these cases is likely to produce even more false positives in different languages, which will be laborious to classify manually.

Identifying street addresses is much more challenging. Addresses may or may not contain digits or punctuation, be expressed as a single string or over multiple HTML tags or JSON fields. Below are some examples of addresses we've encountered in the sampled CC documents:

\begin{verbatim}
Montevideo 938, 3piso - CABA
                        —
<meta itemprop="address" content="4/64-66 Castlereagh St.">
                        —
"address": {
    "@type": "PostalAddress",
    "addressCountry": "RO",
    "addressLocality": "Rosu",
    "addressRegion": "Ilfov",
   "streetAddress": "Nuferilor 16a"
}
                        —
<span itemprop="streetAddress">12417 Woodgreen St</span>
<span itemprop="addressLocality">Los Angeles</span>
<span itemprop="addressRegion">CA</span>
                        —
Pantile House | Newlands Drive | Witham | CM8 2AP | UK
                        —
http://maps.google.ca/maps?...&amp;q=800+Steeles+Ave+W
+Vaughan,+York+Regional+Municipality,+Ontario&amp;...
                        —
1-\u0432\u044b\u0439 
\u0422\u0432\u0435\u0440\u0434\u044b\u0439
\u043f\u0435\u0440\u0435\u0443\u043b\u043e\u043a 7
\end{verbatim}

In addition to varied formats, many texts are in non-Latin scripts, such as Cyrillic, Arabic, Greek, and Chinese/Japanese/Korean characters, occasionally even Unicode-escaped (see final example above). Popular open-source software packages such as \texttt{pyap}\footnote{\href{https://github.com/vladimarius/pyap}{https://github.com/vladimarius/pyap}} or \texttt{libpostal}\footnote{\href{https://github.com/openvenues/libpostal}{https://github.com/openvenues/libpostal}} usually only work with shorter strings and are typically limited to particular countries. In 2010, Chang and Li \cite{chang_mapmarker_2010} proposed MapMarker, a sequence labelling algorithm coupled with pattern mining, to perform postal addresses extraction from web texts, achieving the F-score of 91\%. In 2013, Schmidt et al. \cite{schmidt_extraction_2013} suggested a hybrid approach using patterns and data from gazetteers to extract addresses from web documents, showing high recall and precision. However, in their approach HTML tags are removed from the documents, which is not ideal given that geospatial information can be housed inside \texttt{<meta>} and other tags, as well as inside JavaScript code snippets and JSON objects. In 2018, Efremova et al. \cite{efremova_geo-tagging_2018} introduced a geo-tagging framework that employs n-grams, sliding windows, and a Support Vector Machine classifier to extract addresses from CC documents, achieving precision and recall of 91\% and 93\% respectively. Named-entity recognition tools such as spaCy\footnote{\href{https://spacy.io/}{https://spacy.io/}} can also be trained to extract addresses, yet this would require a labelled training set. 

Inspired by the \textit{needle-in-a-haystack} evaluation task for LLMs\footnote{This popular GitHub repository contains needle-in-a-haystack benchmark tests for LLMs: \href{https://github.com/gkamradt/LLMTest\_NeedleInAHaystack}{https://github.com/gkamradt/LLMTest\_NeedleInAHaystack}}, in which long-context window LLMs are tasked to retrieve particular bits of information from a long document, such as an article or a book, we decided to employ a language model to retrieve both coordinates and addresses from web documents (if they exist), and then manually review the results. LLMs are excellent zero-shot learners and have demonstrated superior performance in named entity recognition and text classification tasks \cite{xie_empirical_2023,zhou_universalner_2024,wei_finetuned_2022,geminiteam2024gemini_manual}, so we believe they are better suited to identify the presence of geospatial in multilingual CC documents than traditional, language-and country-specific algorithms.

\subsection{Gemini 1.5 for geospatial detection}

\begin{figure}[h]
  \includegraphics[width=8cm]{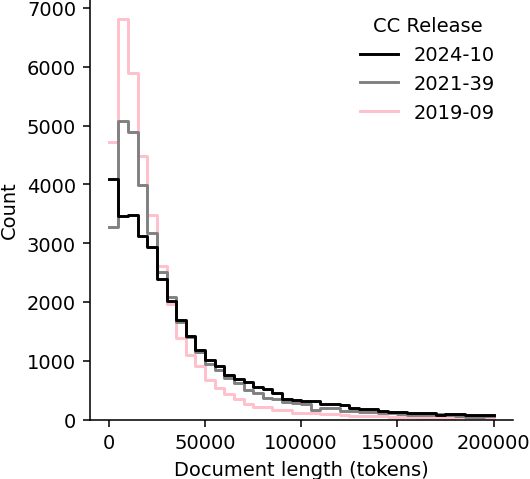}
  \caption{The distribution of sampled web documents' length, in tokens. We capped the x-axis at 200,000; some documents reach 1M tokens in length.}
  \Description{Web document length in tokens (includes query text)}
  \label{token_hist}
\end{figure}

Gemini 1.5 is one of the few LLMs currently capable of handling inputs of up to 1 million tokens (sub-word units), and has shown to perform well in the \textit{needle-in-a-haystack} retrieval task \cite{geminiteam2024gemini_manual}. The context window consideration is important given the web documents found in CC can be lengthy. Figure \ref{token_hist} shows the histogram of sampled document lengths (in tokens) for the documents used in our analysis. While an average document in CC, depending on the release, is between 20-50k tokens, the length distribution is heavily skewed to the right, with some web documents being as long as 1M tokens.

In this paper, we use Gemini 1.5 Flash, which is a lighter and faster version of Gemini 1.5 that also has the most generous free tier API usage allowance of 1,500 requests per day. At the time of running our experiment, the \textit{pay-as-you-go} tier with up to 10,000 requests per day has been announced, but is not yet available\footnote{According to \href{https://ai.google.dev/pricing}{https://ai.google.dev/pricing}, the pay-as-you-go pricing for Gemini 1.5 Flash and Pro will become available on May 30, 2024}.

After some instruction-tuning on sampled web documents from CC, we adopted the following query:

\vspace{0.5cm}
\shadowbox{%
    \begin{minipage}{7.2cm}
Document: ```\texttt{\{cc\_web\_document\}}'''

Question: Does the document enclosed in triple quotes
contain geographic coordinates, such as a latitude-
longitude pair, or a street address that can be used
to identify a physical location such as a building?
If so, return a few examples. If not, just say `None'.
    \end{minipage}
}
\vspace{0.02\linewidth}

We assume the web documents do not contain adverse instructions that may confuse the language model.

Initially, we wanted to use separate queries for coordinates and street addresses; however, due to the API limitations, we decided to combine both elements into a single query to half the execution time after confirming that Gemini responds with both categories of examples (coordinates, addresses) where both are present in the document.

The examples of Gemini responses for queries with documents containing (1) both an address and coordinates, (2) coordinates only, and (3) address only are shown below:

\vspace{0.5cm}
\fbox{%
    \begin{minipage}{7.2cm}
    The document contains the following geographic information:

    * **Latitude-Longitude Pair:** 45.3702375; -85.0128508 
    
    * **Street Address:** 1889 Petoskey Harbor Road
    
    * **City, State, Zip Code:** Petoskey, MI 49770
    \end{minipage}
}

\vspace{0.5cm}
\fbox{%
    \begin{minipage}{7.2cm}
    The document contains geographic coordinates:

     * **Latitude:** -7.3860078
     * **Longitude:** 34.9562661 

     These coordinates are found within the `<div class="single\_job\_listing">` element.
    \end{minipage}
}

\vspace{0.5cm}
\fbox{%
    \begin{minipage}{7.2cm}
    ```
     SCIENCE MUSEUM
     EXHIBITION ROAD
     SOUTH KENSINGTON
     LONDON SW7 2DD
     ```
     
     ```
     https://www.google.co.uk/maps/dir//Science+Museum, +Exhibition+Rd,+Kensington,+London+SW7+2DD
     ```
    \end{minipage}
}

\subsection{Determining sample size}

Previous studies that analysed CC have done so on many randomly sampled documents (often an arbitrary, but large and round number). For example, Luccioni and Viviano \cite{luccioni_whats_2021} sampled 1\% of the English language documents in a single CC release, which amounted to just under 6 million documents. Dodge et al. \cite{dodge_documenting_2021} randomly sampled and clustered 100,000 documents from CC that were block-listed by (and hence did not make it into) the C4 dataset.

However, in our task we are unable to utilise fast algorithms such as keyword lookup, n-grams, or K-Means. We are limited by the number of API requests we can make; in addition, each query takes Gemini between several seconds to several minutes to process. To choose the smallest yet scientifically sound sample size, we settled on Cochran's formula (\ref{eq:1}) \cite{bartlett_organizational_2001}. The formula is used to calculate the ideal sample size that estimates population characteristics with a desired level of precision and a particular confidence interval. It is commonly used in social sciences to estimate the number of people to survey. We use the version of Cochran's formula modified for very large (or infinite) population because a single CC release contains billions of web pages:

\begin{equation} \label{eq:1}
    x = \frac{Z^2 \times p \times (1-p)}{e^2},
\end{equation}

where $e$ is the desired level of precision (we chose 0.005 to get the estimate within $\pm 0.5\%$), $p$ is the estimated proportion of the population that possesses a characteristic (in our case, this is what we are trying to determine, so we leave it at $0.5$ as is customary in such situations), and $Z$ is the z-value of our chosen confidence interval ($\approx1.96$ for 95\% confidence).

Therefore, according to the formula, we need to sample 38,416 documents to get a prevalence estimate approximated at $\pm 0.5\%$ with a confidence interval of 95\%. We decided to sample 39,900 documents (133 random documents from each of 300 parts) from each of 3 individual CC releases. The headroom of 1,400+ documents is to account for the fact that Gemini occasionally fails to process queries, either due to an \emph{Internal error} (code 500) or by marking the query as \emph{Unsafe}. At the end, we queried the language model $3 \times 133 \times 300 = 119,700$ times, and then manually reviewed all responses.

\subsection{Sampled documents}

To check that our sampling is representative of the studied releases, we compare the frequency of primary document languages between our sample and the releases as reported by Common Crawl \cite{common_crawl_statistics_2024}. The primarily language is determined by CLD2\footnote{https://github.com/CLD2Owners/cld2}.

\begin{figure}[h]
  \includegraphics[width=7.5cm]{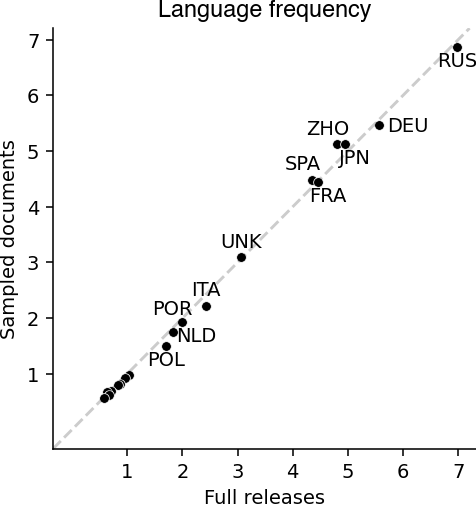}
  \caption{Language frequency among sampled documents vs full releases for the 20 most common languages excluding English (which, at 45\%, dwarfs all other languages). \emph{UNK} is for unknown languages.}
  \Description{Language frequency distribution}
  \label{lang_scatter}
\end{figure}

\begin{figure*}[h!]
  \includegraphics[width=\textwidth]{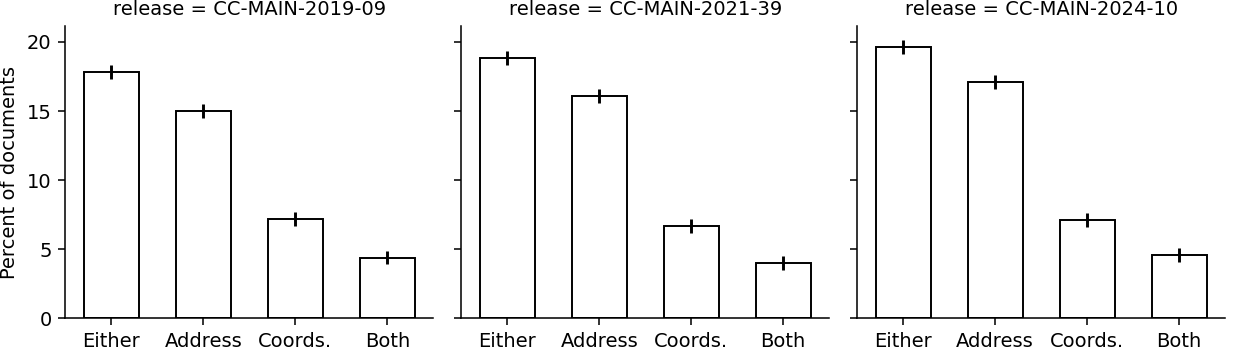}
  \caption{The prevalence of geospatial data in select CC releases, estimated within $\pm0.5\%$ at $95\%$ confidence }
  \Description{The prevalence of geospatial data in CC documents}
  \label{result}
\end{figure*}

English is the most common primary language of web documents in CC, representing 45.0\% of the sampled releases \cite{common_crawl_statistics_2024} and 45.6\% of our sample. The frequency of the other most common 19 languages is shown in Figure \ref{lang_scatter}. We can see that our sample is balanced in relation to most common CC languages. It is worth noting that the frequency does not reflect the real-world popularity of languages \cite{paolillo_measuring_2005}. This is partly explained by the fact that the Internet originated and was popular in the Anglo-sphere for longer than the rest of the world, and because CC crawling infrastructure is located in the United States, which may force websites to serve English-language pages by default \cite{baack_critical_2024}.

\begin{table}
  \caption{Most frequent ($>1\%$) TLDs of sampled documents}
  \label{tab:tlds}
  \begin{tabular}{llr}
    \toprule
     TLD & Country & Frequency (\%) \\
    \midrule
    .com & ---  &  45.4 \\
    .org & ---  &   6.0 \\
    .ru  & Russia  &   4.6 \\
    .de  & Germany  &   3.9 \\
    .net & ---  &   3.7 \\
    .uk  & United Kingdom  &   2.3 \\
    .jp  & Japan  &   2.1 \\
    .fr  & France   &   1.6 \\
    .it  & Italy  &   1.6 \\
    .nl  & Netherlands  &   1.5 \\
    .pl  & Poland  &   1.4 \\
    .br  & Brazil  &   1.2 \\
    .cn  & China  &   1.1 \\
    .au  & Australia  &   1.1 \\
    .edu & --- &   1.1 \\
  \bottomrule
\end{tabular}
\end{table}

Table \ref{tab:tlds} lists the most frequent top-level domains (TLDs) of the sampled documents. These are largely consistent with the summary provided by CC \cite{common_crawl_statistics_2024} for full releases, and the Internet-wide domain survey by W3Techs \cite{w3techs_usage_2024}. The single most frequent TLD is \emph{.com}, which accounts for over 45\% of all sampled documents. While \emph{.com} is a generic TLD that can be used by any person, organisation, or business anywhere in the world, it is especially popular in the United States in place of the country-specific \emph{.us} domain.

\subsection{Reviewing Gemini responses}

We manually reviewed each Gemini response that is longer than 1 or 2 tokens (for \emph{`None'} and \emph{`None.'}, respectively) in order to check that we are satisfied with the quality of the examples returned, as well as to distinguish whether the document contains an address, or coordinates, or both.

To ensure Gemini did not hallucinate the examples, that is, made up random coordinates or addresses that do not exist in the source document, we manually checked a random sample of 100 documents for which the language model returned geospatial examples. We were able to locate \emph{all} examples provided by Gemini in the corresponding source documents.

In order to check for false negatives, that is, the documents for which Gemini returned \emph{None} that may contain geospatial data, we manually reviewed a random sample of 100 such documents. Using Google Translate\footnote{https://translate.google.com} when needed, we were unable to identify street addresses or coordinates in \emph{all} cases.

\section{Findings}

Our estimate of the prevalence of geospatial data in CC documents is shown in Figure \ref{result}. Overall, when analysing the three CC releases from between 2019 and 2024, we find that 18.7\% of the documents contain either a street address or a pair of coordinates (or both), 16.1\% of the documents contain a street address, 7.0\% of the document contain a pair of coordinates, and 4.3\% of the documents have both an address and a pair of coordinates. These estimates are fairly consistent between the releases.

Around 27.1\% of the web documents that contain an address also contain a pair of coordinates, and 62.0\% of the documents that contain coordinates also contain an address. (It is worth noting that the coordinate and the address found in the same document do not necessarily refer to the same location.)

\begin{figure}[h]
  \includegraphics[width=8cm]{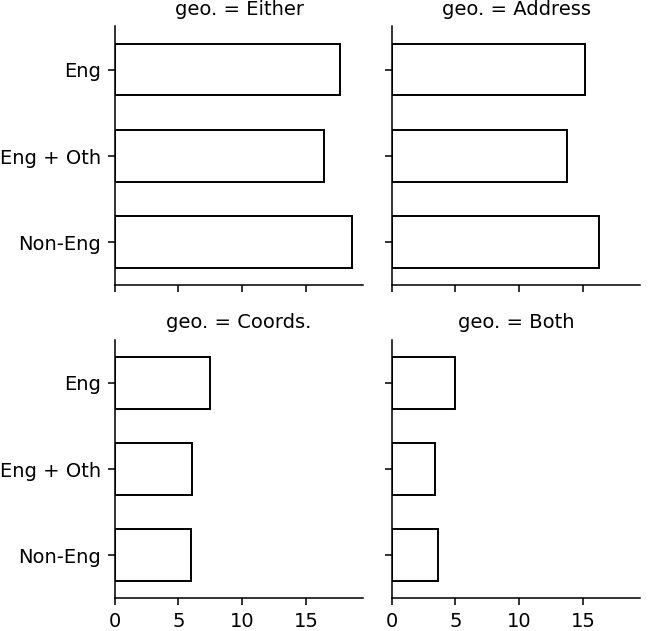}
  \caption{Geospatial data frequency in English and non-English documents. \emph{Eng + Oth} refers to documents that are both in English and some other language.}
  \Description{Geospatial data occurrence by language}
  \label{lang_bar}
\end{figure}

Figure \ref{lang_bar} demonstrates that the frequency of geospatial data occurrence is fairly consistent between documents in English and documents in other languages. While English-language documents are slightly more likely to contain coordinates than non-English-language documents (7.5\% vs 6.0\%), the latter are slightly more likely to contain addresses (16.2\% vs 15.1\%).

The most pronounced single source of coordinates in sampled web documents is URLs to Google Maps, present either as a hyperlink to the mapping service itself (\texttt{<a href="...">}), or as an embedded map (\texttt{<iframe src="...">}). An example of such URL can be found in Subsection \ref{IdentifyingCoordinatesAndAddresses}. Thus, the fact that Google Maps is not widely used in some major non-English speaking countries such as China (where Baidu Maps\footnote{https://map.baidu.com/} and AutoNavi\footnote{https://mobile.amap.com/} are popular), Russia (Yandex Maps\footnote{https://yandex.ru/maps}), Iran or Vietnam, is likely to have a negative impact on the frequency of coordinates in non-English-language web documents.

On several occasions, Gemini judged that an IP address or a phone number or a named landmark (such as a particular theatre) is sufficient to locate the place. In some cases, it also returned the astronomical coordinates of celestial bodies. However, these did not meet our definition of geospatial described in Subsection \ref{WhatConstituesGeospatial}, so we excluded them from the final count.

\section{Conclusion}

In this paper, we analysed the contents of CC for the presence of geospatial data. Using Gemini 1.5 Flash and manual verification on a sample of 119k documents, we estimated with 95\% confidence that $18.7\% \pm 0.5\%$ of web documents in recent CC releases contain a street address or a pair of coordinates. Thus, we conclude that geospatial data is fairly well represented in mostly unstructured CC documents.

Therefore, it is not surprising that LLMs exhibit some geospatial capabilities, such as the knowledge of distances and directions between landmarks and other locations. We showed that documents in the English language and from the \emph{.com} TLD are disproportionately present in CC; this imbalance risks amplifying some worldviews while silencing others, exposing the language models to geospatial biases during pre-training. Therefore, we suggest the need for further research into the accuracy and quality of the geospatial data for different regions and languages, which can help us better understand the geospatial biases that LLMs exhibit. 

We also recognise that coordinates and addresses, being specific point-level bits of information, are not the only way LLMs learn the physical space; the context of the web pages where geospatial data is found is also very important. This may include textual descriptions of places, both objective and subjective, such as those describing the surroundings or the experience of the visitors or the activities happening in the area.

With roughly $250B \times 18\% \approx 46B$ documents potentially containing geospatial data, Common Crawl a potential goldmine for domain-specific datasets that can be used to train, fine-tune, or benchmark LLMs on geospatial tasks.

\begin{acks}
This work was supported by the Ordnance Survey, and the Engineering and Physical Sciences Research Council [grant no. EP/Y528651/1]. The authors would also like to thank Chuang Liu, Zhengxiang Shi, Meihui Wang, Xiaowei Gao, Jerome Ramos, Hao Chen, Xinglei Wang, and Junyuan Liu of UCL SpaceTimeLab for their help facilitating access to the Gemini API.
\end{acks}

\bibliographystyle{ACM-Reference-Format}
\bibliography{articles,articles_manual}


\end{document}